\documentclass[10pt,twocolumn,letterpaper]{article}

\usepackage{wacv}
\usepackage{times}
\usepackage{epsfig}
\usepackage{graphicx}
\usepackage{amsmath}
\usepackage{amssymb}
\usepackage{multirow}
\usepackage{microtype}
\usepackage{graphicx}
\usepackage{subfigure}
\usepackage{booktabs} 
\usepackage{amsfonts}
\usepackage{algorithm}
\usepackage{algorithmic}

\wacvfinalcopy 

\def\wacvPaperID{****} 

\ifwacvfinal\fi
\begin{document}

\title{A ``Network Pruning Network'' Approach to Deep Model Compression}

\author{
Vinay Kumar Verma\hspace{1.2cm}Pravendra Singh\hspace{1.2cm}Vinay P. Namboodiri\hspace{1.2cm}Piyush Rai\\
Department of Computer Science and Engineering, IIT Kanpur, India\\
{\tt\small \{vkverma, psingh, vinaypn, rpiyush\}@iitk.ac.in}
}

\maketitle
\ifwacvfinal\thispagestyle{empty}\fi

    \begin{abstract}
        We present a filter pruning approach for deep model compression, using a multitask network. Our approach is based on learning a a pruner network to prune a pre-trained target network. The pruner is essentially a multitask deep neural network with binary outputs that help identify the filters from each layer of the original network that do not have any significant contribution to the model and can therefore be pruned. The pruner network has the same architecture as the original network except that it has a multitask/multi-output last layer containing binary-valued outputs (one per filter), which indicate which filters have to be pruned. The pruner's goal is to minimize the number of filters from the original network by assigning zero weights to the corresponding output feature-maps. In contrast to most of the existing methods, instead of relying on iterative pruning, our approach can prune the network (original network) in one go and, moreover, does not require specifying the degree of pruning for each layer (and can learn it instead). The compressed model produced by our approach is generic and does not need any special hardware/software support. Moreover, augmenting with other methods such as knowledge distillation, quantization, and connection pruning can increase the degree of compression for the proposed approach. We show the efficacy of our proposed approach for classification and object detection tasks.

    \end{abstract}
     \vspace{-2em}
    \section{Introduction}
    Recent advances in deep learning have led to an impressive and significant breakthroughs in various domains, such as computer vision \cite{resnet,vgg2014very,reed2018fewshot,finn2018probabilistic,verma2018generalized,huang2017densenet,verma2019meta}, NLP \cite{xu2015show,mikolov2013efficient,antol2015vqa}, and information retrieval \cite{liu2017deep}. Pushing the performance further typically leads to models with overly complex, deeper architecture, which tends to increase the model size (number of parameters, depth, and breadth of layers), and FLOPs enormously, and such complex models may not be ideal to be deployed on resource-constrained devices.
    
    This had led to considerable interest in making the model more efficient, in terms of storage as well as computation~\cite{ding2018auto,channelPruning17,singh2018leveraging,li2016pruning,singh2019multi,singh2019falf,mazumder2019cpwc,singh2019play,he2018soft,singh2020cooperative,singh2019accuracy,yu2017nisp}. A popular approach to increase the efficiency of the model is via model compression. Among the existing model compression approaches, the filter pruning based approaches usually show superior performance regarding FLOPs and runtime memory compression \cite{channelPruning17,yu2017nisp,ding2018auto}. 
    
    Selecting the most optimal subset of filters to prune from a Convolutional Neural Network (CNN) model is a combinatorially hard problem. Therefore, the existing filter pruning approaches are based on some heuristics to define filter importances. Recent works~\cite{li2016pruning,channelPruning17} have shown that the strength of the feature map (output produced by a convolutional filter) dominates the output of the network. Filters with a feature map have minimal contribution to the final decision of the model; therefore, the corresponding filters can be removed from the network. In these approaches, the objective is to find the filters that are likely to produce zero (or near-zero) feature map\footnote{A feature map is said to be zero feature map if its $\ell_1$ norm is zero, i.e., all of its elements are zero.}. In a pre-trained model, it is rare to get zero feature map. Therefore we optimize the network such that a majority of the filters (which are going to be pruned) have their feature map value close to zero, while the rest of the filters (that remain in the model) can still achieve accuracy close to the original network. Therefore after discarding the filters that produce zero feature-maps, the model does not incur any significant performance drop.
    
Most of the existing filter pruning approaches are based on heuristics to define filter importance. Defining filter importance is itself a challenging task. Also, before discarding the less important filters from the model, the representational capacity of the less important filters should be transferred to the remaining part of the model. This is a challenging task, and most of the previous approaches \cite{lin2017feature,he2018soft,yu2017nisp} exhibit poor performance in doing so and, consequently, these approaches exhibit a sharp drop in accuracy after a moderate pruning and require a high degree of finetuning, which can be very time consuming in practice. 

Another drawback in the previous approaches \cite{ding2018auto,luo2018thinet,DPruner,Smallify} is that they are unable to decide the layer importance. Performance of a CNN model may be very sensitive w.r.t. some of the layers, and we cannot remove a large number of filters from such layers. In contrast, some other layers may have a high degree of filter-level redundancy. It is a very challenging task to define layer importance precisely. Most previous methods \cite{ding2018auto,channelPruning17,li2016pruning,he2018soft,yu2017nisp} consider this as a hyperparameter (i.e., how many filters to prune from each layer). Therefore, these approaches take as input the number of filters to be pruned from each layer. To set these hyperparameters is an arduous task since, for $K$ layers, we have $K$ such hyperparameters. Therefore it is desirable to develop an automatic method that decides \textit{where to prune} in the model, which motivates our approach.

We present a ``network pruning network'' approach for deep model compression in which we learn a \emph{pruner network} that prunes a target (main) network. The pruner network is essentially a deep multitask network that adaptively decides which filters to prune in each layer of the target network. The objective of the multitask network is to learn weights corresponding to each output feature map of the main network (which we are going to prune) such that most of the feature maps are zero weighted without sacrificing the accuracy. Therefore the filters that correspond to zero feature maps can be safely removed from the main network without hurting the main network performance. In the proposed approach, the multitask network contains the same CNN architecture as the main network (e.g., ResNet for ResNet) but contains task-specific output layers consisting of binary outputs that denote the filters that have close to zero feature maps. The multitask network learns to maximize the number of zero feature maps in the main network. The proposed approach is end-to-end trainable using gradient descent. Our main contributions can be summarized as follows:
\vspace{-0.6em}
\begin{itemize}
    \item The proposed approach leverages the idea of \emph{multitask-learning}, which guides us on how to prune in each layer. We can obtain a compressed model using just a few epochs without any significant accuracy drop.
    \vspace{-0.6em}
    \item The proposed approach uses a \emph{multitask} network, which adaptively learns the filter importance in an end-to-end trainable manner in contrast to existing filter pruning approaches that rely on \emph{ad hoc} heuristics to calculate the filter importances.
    \vspace{-0.6em}
    \item Most of the existing approaches \cite{ding2018auto,channelPruning17,li2016pruning,he2018soft,yu2017nisp} require specifying \textit{how many filters from each layer to prune} or require \textit{a threshold} that is used to determine which filters to prune. In the proposed approach, we do not require any such input and can automatically learn the \emph{layers importance}, thereby reducing the number of hyperparameters.
\end{itemize}

Note that, although our approach consists of two networks, i.e., a deep network to prune another deep network, it is different from student-teacher based knowledge distillation approaches~\cite{hinton2015distilling} to deep model compression where the idea is to compress a teacher network into a simpler student network. In contrast, our approach learns a deep multitask network that prunes a target network. 

    \begin{figure*}[t]
        \centering
        \includegraphics[scale=0.55]{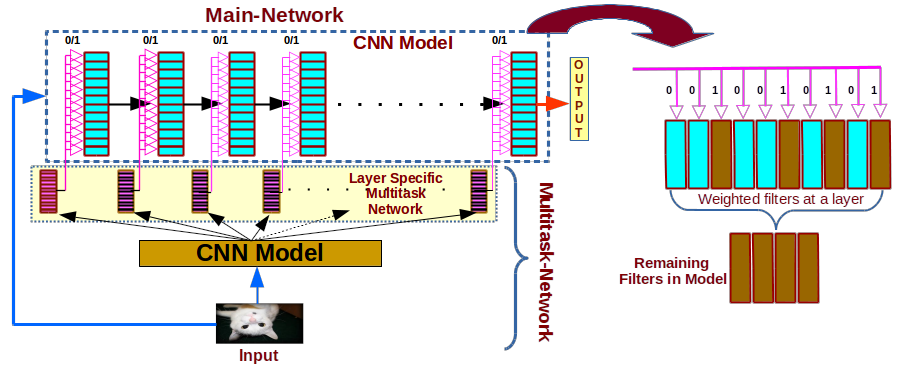}
        \caption{The upper architecture is the main network that we wish to prune, and the lower model is the same as the original one, with the only difference being that the multitask architecture replaces the output layer. Each has a task to prune a layer in the main network.}
        \label{fig:complete_model}
        \vspace{-1.4em}
    \end{figure*}
    \vspace{-1em}
    \section{Proposed Approach}

    \subsection{Notation} \vspace{-0.7em}
    Let us assume a CNN architecture with $K$ convolutional layers.
    Assume $\mathcal{L}_i$ to be the $i^{th}$ layer and $i\in [1,2,\dots K]$. The layer  $\mathcal{L}_i$  has $n_i$ filters which gives the $n_i$ feature-maps that are used as input for the next layer. The set of filters at layer $\mathcal{L}_i$ is denoted as $\mathcal{F}_{\mathcal{L}_i}$ where $\mathcal{F}_{\mathcal{L}_i}=[f_1,f_2,\dots, f_{n_i}]$. Similarly, the feature maps at layer $\mathcal{L}_i$ are represented as $\mathcal{M}_{\mathcal{L}_i}=[m_1,m_2,\dots, m_{n_i}]$. Each feature map $m_i$ is of dimension $(h_k,w_k)$, where $h_k$, $w_k$ are height and width, respectively, of the feature map. Therefore the shape of $\mathcal{M}_{\mathcal{L}_i}$ is $(h_k,w_k,n_i)$.
    
    \vspace{-0.7em}
    \subsection{Model}
    \vspace{-0.3em}
    This section briefly describes how the multitask network is used to prune the filters from the main network (the CNN). The core idea of our approach is to design a \emph{multitask network} that \emph{learns} a weight for each filter in the main network, and optimizes the main network such that most of the filters produce zero feature maps after being weighted by the multitask network. Corresponding filters in the main network that produce zero feature maps do not have any significant contribution to the model performance and can be discarded from the main network without sacrificing the model's performance.   
    
    Our approach is based on learning the weights for each filter in the main network. However, instead of associating weights to each filter, we associate weights with each feature map (the output produced by a filter of the main network). The multitask network learns these weights. The objective of the multitask network is to maximize the number of zero weights corresponding to output feature maps in the main network.  The multitask network has the same architecture as the main network that we would like to prune (e.g., for pruning the ResNet main network, the multitask network is also ResNet architecture with a modified output layer). Essentially, to prune a model with $K$ layers, we have a multitask network with $K$ outputs, where the $K$ outputs themselves have dimensions of size $[n_1,n_2, \dots, n_K]$. Here $n_i$ is the number of filters at layer $\mathcal{L}_i$. We refer the \emph{main network} as ($\mathcal{O}$) while the multitask network is called the \emph{pruner} ($\mathcal{P}$). Fig.~\ref{fig:complete_model} summarizes the complete architecture of our proposed model compression framework.
    
    Suppose the main network ($\mathcal{O}$) has a cost function $C_O(\Theta_o)$, where $\Theta_o$ denotes the parameters of the network $\mathcal{O}$. Also assume that the pruner network ($\mathcal{P}$) has a cost function $C_P(\Theta_m)$, where $\Theta_m$ denotes the parameters of the network $\mathcal{P}$. The architecture of the \emph{pruner} is the same as the \emph{main-network} ($\mathcal{O})$; the only difference is that the output layer is replaced by a  multitask network that has $K$ outputs (number of layers in the $\mathcal{O}$), with each of the $K$ outputs itself being a \emph{binary vector}. The size of the vector at layer $\mathcal{L}_i$ is $n_i$ (size equal to the number of filters at $\mathcal{L}_i$). The complete model is shown in Fig.~\ref{fig:complete_model}.
    \vspace{-0.6em}
    \subsection{Main-Network ($\mathcal{O}$)}
    
    The \emph{main network} corresponds to the original network that we would like to prune. The only difference from the original network is that feature maps $\mathcal{M}_{\mathcal{L}_i}$ on each layer $\mathcal{L}_i$ are replaced by \emph{weighted} feature maps, and the weights are given by the \emph{pruner network} ($\mathcal{P}$) (explained in the next section). Let $\mathbf{W}_{\mathcal{L}_i}=[w_1,w_2,\dots, w_{n_i}]$ be the weights of layer $\mathcal{L}_i$ given by the $\mathcal{P}$ network. Then $\mathcal{O}$'s $\mathcal{L}_i^{th}$  layer feature maps are replaced as:
    \vspace{-0.7em}
    \begin{equation}
    \label{eq:weighted_featuremap}
    \mathcal{M}_{w}=[w_1m_1,w_2m_2,\dots, w_{n_i}m_{n_i}]
    \end{equation}
    Here $m_1,m_2,\dots m_{n_i}$ are the feature maps at layer $\mathcal{L}_i$ in original network. 
    Now our objective is to optimize the network with the help of $\mathcal{P}$ such that most of the $w_i$'s are close to zero, without sacrificing the accuracy. The complete objective and joint loss are described in Section \ref{sec:complete_model}.  The modified network can be easily optimized with the help of any gradient descent based optimizer.
    
    Therefore, in the complete network, we represent each feature map as $\tilde{m}_j = w_jm_j$, here $w_j\in[0,1]$ i.e. each feature map $m_j$ is weighted by a weight $w_j$. The \emph{pruner network} learns these weights. In the main network, $\forall w_j: w_j\rightarrow 0$ does not have any significant contribution to the overall network performance, implying that $m_j$ can be pruned from the model.
    
    Therefore, by discarding all the filters $f_i$ corresponding to the $w_i \approx 0$ (feature-maps weights) from the network $\mathcal{O}$, do not significantly degrade the model's performance. Hence we can remove all such filters and corresponding feature maps from the model.
    \vspace{-0.5em}
    \subsection{Pruner Network ($\mathcal{P}$)}
\begin{figure}[!t]
        \centering
        \includegraphics[scale=0.55]{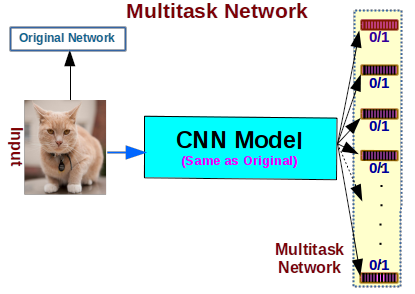}
        \caption{\textbf{Multitask pruner network:} Multitask Network has the same base model as main network but the output later is replaced by layer specific multitask network.}
        \label{fig:meta-learner}
        \vspace{-1.4em}
    \end{figure}

    The \textit{pruner network} is the network that is responsible for the filter pruning in the main network. The \emph{pruner network} maximizes the number of zero feature-maps in the main network. The corresponding filters that produce the zero feature-maps can be discarded from the main network. The \emph{pruner network} give weights to each feature-maps in the main network and tries to optimize weights such that most of the $w_i\rightarrow 0$. Our \emph{pruner network} $\mathcal{P}$ is a multitask network, with the base network being the same as the \emph{main-network}, and the fully connected output layer replaced by multitask output layers. The number of output in multitask output layers is same as the number of layers in the model $\mathcal{O}$, i.e., $K$. The dimension of each multitask output is $n_i$ (number of filters on layer $\mathcal{L}_i$). The \emph{pruner} multitask network is shown in Fig.~\ref{fig:meta-learner}.
    
    Let's assume that the \emph{pruner network} has the cost function $C_P(\Theta_m)$, where $\Theta_m$ denotes its parameters. We need to optimize the model such that each of the outputs in the multitask output layer is binary, i.e., $\forall w_i: w_i \in \{0,1\}$. To retain differentiability, we approximate the Bernoulli outputs using a scaled sigmoid on the output values. This scaled sigmoid gives a sharp change between 0 and 1. A moderate-scale value of the sigmoid can approximate the Bernoulli distribution. The scale of the sigmoid is increased gradually since experimentally we found that, if initially, we set high scale value in the network then the network is unable to learn. 
    
    Let ${f(\Theta_m)}$ be the output of the network $\mathcal{P}$. i.e.:
    \begin{equation}\label{eq:pruner}
    \small
    f(\Theta_m)=\mathbf{[\mathbf{W}_{\mathcal{L}_1},\mathbf{W}_{\mathcal{L}_2},\dots,\mathbf{W}_{\mathcal{L}_K}]} \quad \forall \mathbf{W}_{\mathcal{L}_i} \in[0,1]^{n_i}
    \end{equation}
    Here $\mathbf{W}_{\mathcal{L}_i}$ denotes the $i^{th}$  output of the multitask network and is of size $n_i$ (number of filters on layer $\mathcal{L}_i$). In the next section, we briefly explain how we can achieve the objective of Eq.~\ref{eq:pruner} without affecting the model performance. Eq.~\ref{eq:pruner} gives the weights to each feature maps on every layer. The output that has the zero value gives the zero weight to the corresponding feature-map, and we can discard this feature-map and corresponding filter from the \emph{main-network} without degrading the model performance. The objective of this network is to maximize the number of zeros in the multitask output space. The alternate optimization of the $\mathcal{O}$ and $\mathcal{P}$ ensure that the accuracy drop is minimal in the filter pruning process. In the first round, only $\mathcal{P}$ is optimized, while the parameters of $\mathcal{O}$ are kept frozen. In the second round, $\mathcal{P}$ and $\mathcal{O}$ is optimized jointly. Optimizing $\mathcal{P}$ tries to minimize the number of filters/feature-maps in the main network while optimizing $\mathcal{O}$ recovers the accuracy. 
        \begin{algorithm}[!t]
            \caption{Multitask Network for Model Compression}
            \label{alg:maml}
            \begin{algorithmic}[1]
                \REQUIRE $C_{MP}(\Theta_o,\Theta_m)$: The complete model
                \REQUIRE $\alpha$ and $\beta$: learning rate and $N$: \#epoch
                \STATE Initialize $\Theta_o$ and $\Theta_m$ from pretrained model
                \WHILE{epoch$\leq$ N}
                \IF{epoch\%2==0}
                \small
                \STATE Calculate $\Theta_m'$ by Eq:\ref{eq:metalearner}\\
                i.e. $\Theta_m' \leftarrow \Theta_m-\alpha\bigtriangledown_{\Theta_m} (C_{MP}^{t}(\Theta_o,\Theta_m)+\lambda||f(\Theta_m)||_{l_1})$
                \ELSE
                \STATE Update $[\Theta_o,\Theta_m]$ using latest value $[\Theta_o,\Theta_m']$ by Eq:\ref{eq:main_network} i.e.
                $[\Theta_o,\Theta_m] \leftarrow [\Theta_o,\Theta_m']- \beta\bigtriangledown_{[\Theta_o,\Theta_m']} ( C_{MP}^{t+1}(\Theta_o,\Theta_m') +\lambda||f(\Theta_m')||_{l_1})$
                \ENDIF
                \ENDWHILE
                \STATE Remove all the filters and corresponding feature maps having $w\rightarrow 0$ from the main network ($\mathcal{O}$) 
                \STATE Finetune the pruned model with the remaining filters
            \end{algorithmic}
        \end{algorithm}    
    Notably, our proposed approach essentially transforms the model compression problem as an end-to-end optimization problem. This can be easily optimized using stochastic gradient descent (SGD). The proposed approach automatically select the filters from each layer based on the layer importance. This fact can be easily verified since in our final compressed model's different layers have different compression rates. In contrast, other existing approaches \cite{channelPruning17,li2016pruning,he2018soft,yu2017nisp} need the number of filters to be pruned from each layer as the hyperparameters. The multitask pruner network is shown in  Fig.~\ref{fig:meta-learner}.

    \textbf{Producing the Binary Weights:} Generating the binary weights on the multitask output layer is a key point that controls the pruning rate. A high zero cardinality results in a high pruning at the cost of accuracy drop, while low zero cardinality produces a low pruning. We have to make a trade-off between the number of zeros and the accuracy drop. To produce values close to 0/1, we adopted the scaled sigmoid on the multitask output space, along with $l_1$ regularizer. The $l_1$ regularizer produces the sparsity on the output space, and sparsity can be controlled by the regularization constant. Initially, we set the scale as 1, and after a few epochs, we changed it to 30; it helps to convert the sigmoid function to nearly a step function for 0/1. The main advantage of the scaled sigmoid is that it is differentiable. The more details for each architecture are given in the experiments section.
    
    \subsection{The Complete Model}
    \label{sec:complete_model}
    This section explains how the complete objective is defined and the optimization is performed over the \emph{main network} $\mathcal{O}$ and \emph{pruner network} $\mathcal{P}$. The section also explains how the \emph{multitask network} learns to prune the \emph{main-network}. 
    
    Let $C_{MP}$ be the joint loss of the \emph{main network} and the \emph{pruner network}, and $C_O(\Theta_o)$ and $C_M(\Theta_m)$ be the cost functions defined for $\mathcal{O}$ and $\mathcal{P}$, respectively. The joint objective can be defined as:
    \begin{equation}
    \vspace{-0.5em}
    \label{eq:joint}
    \min_{\Theta_o} \min_{\Theta_m} C_{MP}(\Theta_o,\Theta_m)+\lambda||f(\Theta_m)||_1  
    \end{equation}
    Here $C_{MP}$ is the joint loss w.r.t parameters $\Theta_o$ and $\Theta_m$. $\lambda$ is the regularization constant and $||f(\Theta_m)||_1$ is the $\ell_1$ penalty on the output of the pruner network. The epoch $t$ updates for the pruner network is given by
    \begin{equation}
    \small
    \label{eq:metalearner}
    \Theta_m' \leftarrow \Theta_m-\alpha\bigtriangledown_{\Theta_m} \left(C_{MP}^{t}(\Theta_o,\Theta_m)+\lambda||f(\Theta_m)||_1\right)
    \end{equation}
    In Eq. \ref{eq:metalearner}, the gradients are calculated only w.r.t. \emph{pruner network} parameters $\Theta_m$. The optimal parameters for pruner network  $\Theta_m'$ are obtained in epoch $t$ will be used as input in the next epoch ($t+1$) to train the main-network. The optimization of the main network can be given as:
    \begin{equation}
    \label{eq:main_network}
    \begin{split}
    [\Theta_o,\Theta_m] \leftarrow [\Theta_o,\Theta_m']- & \beta\bigtriangledown_{[\Theta_o,\Theta_m']} ( C_{MP}^{t+1}(\Theta_o,\Theta_m') \\ +\lambda||f(\Theta_m')||_{l_1})
    \end{split}
    \end{equation}
    
    Here $[\Theta_o,\Theta_m]$ denotes the joint parameters of both models  and Eq.~\ref{eq:main_network} uses the most recent values  $\Theta_m'$ of the optimal parameters of the \emph{pruner network}. $\alpha$ and $\beta$ are the learning rate for Eq.~\ref{eq:metalearner} and \ref{eq:main_network}, respectively. The optimization of Eq.~\ref{eq:main_network} is performed jointly w.r.t. parameters $\Theta_o$ and $\Theta_m$.
    The optimization in Eq.~\ref{eq:metalearner} maximizes the number of zeros in the output of the \emph{pruner network} because of $\ell_1$ penalty. At the same time, it also minimizes the loss; therefore, the model performance is maintained. Eq.~\ref{eq:main_network} optimizes the model w.r.t. all parameters. Therefore, the main network has the flexibility to transfer representational capacity of the less important filters to the remaining part of the model (so as to maintain the representational capacity).
    
    It is interesting to note that the two-step update defined by Eq.~\ref{eq:metalearner} and \ref{eq:main_network} is akin to the updates of a model-agnostic meta-learning (MAML) framework~\cite{maml}. The only difference is that, in MAML, the optimization of \emph{meta-learner} and \emph{main-learner} are done over the same set of parameters. In contrast, in our proposed model compression approach, the \emph{pruner network} parameters are a subset of the main learner's parameters. Unlike the original MAML~\cite{maml} framework, there is also no ``task distribution'' over a dataset since here the model pruning is a single, stand-alone task that we wish to solve.
    
    \vspace{-0.7em}
    \section{Related Work}
    
    Most of the recent work on model compression can be categorized into three broad categories: connection pruning, filter pruning, and quantization. The filter pruning approach has been more popular as compared to the other methods since it gives the maximum practical speedups and minimizes runtime memory, without requiring special hardware/software support. The other methods usually require special hardware/software support.
    \vspace{-0.7em}
    \subsection{Connection Pruning}
    In deep CNNs, most of the weights are redundant in the model. The connection pruning is a simple method to introduce sparsity in CNN model parameters. It prunes the redundant connections from the model. One approach to compress the CNN architecture is to prune the unimportant/redundant parameters. However, it is challenging to define the importance of the parameters quantitatively. There are many approaches to rank the importance of the parameters. Optimal Brain Damage \cite{lecun1990optimal} and Optimal Brain Surgeon \cite{hassibi1993second} used the second-order Taylor expansion to calculate the parameters importance. These approaches are based on the calculation of the second-order derivative and therefore are very costly. Blockdrop \cite{wu2018blockdrop} proposed the skip layer approach for network compression. Using the hashing function, the method proposed in \cite{chen2015compressing} randomly groups the connection weights into a single bucket and then finetunes the network to recover the performance. \cite{han2015deep} proposes an iterative approach where absolute values of weights below a certain threshold are set to zero, and the drop in accuracy is recovered by finetuning. The connection pruning approach is very successful when most of the parameters lie in the fully connected layer. However, these approaches result in unstructured sparsity in the model. Special hardware/software adds extra overhead. Another disadvantage of these approaches is that they are unable to save the runtime memory GPU memory.
    
    \vspace{-0.7em}
    \subsection{Filter Pruning}
    
    Unlike the connection pruning approach, the filter pruning approach \cite{hu2016network,singh2019stability,li2016pruning,yu2017nisp} discards the whole filter from the model. As a result, the depth of feature maps is also reduced. The filter pruning approaches (which is the focus of our work too) do not need any special hardware or software for acceleration. Filter pruning approaches can be categorized into two categories. One class of methods find out the important filters in the model and discard the unimportant ones. After that, at each pruning step, re-training is needed to recover from the accuracy drop.  \cite{hu2016network} evaluates the importance of filters on a subset of the training data based on the output feature maps.  \cite{abbasi2017structural} used a greedy approach for pruning. They evaluated the filter importance by checking the model accuracy after pruning the filters.  \cite{molchanov2016pruning} and \cite{li2016pruning} used a similar approach but a different metric for filter pruning. The filter pruning approach in \cite{li2016pruning} is mostly based on the weight magnitude of the filters.  \cite{denton2014exploiting,zhang2015efficient,jaderberg2014speeding} used the low-rank approximation which relies on matrix factorization and can thus be costly in practice. \cite{liu2017learning} performed channel-level pruning based on the scaling factor in the training process. The pruning is done layer by layer; hence, it is very slow.
    The group sparsity-based approaches have also become popular for filter pruning. \cite{lebedev2016fast,wen2016learning,zhou2016less,alvarez2016learning} explored the filter pruning based on the group sparsity. 
    
    In the same vein as our work, recently \cite{lin2017runtime,He_2018_ECCV,rao2018runtime} proposed automatic filter pruning approaches. \cite{lin2017runtime,rao2018runtime} proposed a reinforcement learning-based approach; it finds a dynamic routing path at run time to prune the model. In \cite{He_2018_ECCV} another reinforcement learning-based model has proposed that leverage on the actor-critic model for the network pruning. These approaches use a dynamic pruning policy, while in the proposed approach, we use a single policy. Also, the proposed model was not based on reinforcement learning-based algorithms. Another popular approach are the design the efficient CNN model that can train the network from the scratch \cite{zhang2018shufflenet,zhang2018shufflenet,singh2019hetconvijcv,singh2019hetconvijcv}. Our work focus on the filter pruning and the efficient CNN based approach are the out the scope of this work.

    \vspace{-0.7em}
    \subsection{Quantization}
    The method in \cite{han2015deep} compressed the CNN by combining pruning, quantization, and Huffman coding. In \cite{miao2017towards}, the proposed compression method was based on the floating point value quantization for model storage. These approaches assume that 32-bit float representation is redundant for the model parameters. Here we can use a lower bit configuration for model parameters without sacrificing the performance. The extreme case of this approach can be binary bit quantization. Binarization \cite{binarycompression} for model parameters can be used for the network compression where each floating point value is quantized to a binary value. Bayesian methods \cite{louizos2017bayesian} have also been used for the network quantization. Most quantization methods require special hardware support to get the advantage of the compression. 
    
    
    \vspace{-.7em}
    \section{Experiments and Results}
    To show the effectiveness of our proposed approach, we perform extensive experiments on large as well as small-scale datasets. We perform experiments on  ResNet-50 \cite{resnet} and VGG-16 \cite{vgg2014very} architecture over the large-scale dataset ImageNet \cite{imagenet2015}. We also conduct experiments on ResNet-56 \cite{resnet} and VGGLike \cite{vgg2014very} architecture over the small scale dataset CIFAR-10 \cite{cifar10} . To show the generalization ability of our proposed approach, we also conduct an experiment over the large scale MS-COCO \cite{lin2014coco} dataset using Faster-RCNN object detector. Our experimental results demonstrate that the proposed approach yields state-of-art model compression.
    \vspace{-1em}
    \subsection{Implementation Details}
    The proposed framework consists of two networks, \emph{pruner network} $\mathcal{P}$ and \emph{main network} $\mathcal{O}$. The \emph{pruner network} $\mathcal{P}$ gives the weights to the feature maps of \emph{main-network} $\mathcal{O}$. The objective of $\mathcal{P}$ is to maximize the number of zeros in the multitask output space, whereas $\mathcal{O}$ maintains the accuracy drop because of $\mathcal{P}$. The task given to $\mathcal{P}$ is easier than the task given to the $\mathcal{O}$. The \emph{pruner network} can quickly maximize the number of zeros in the output space, but empirically, we find that this gives the sharp accuracy drop in the model. Since the quick optimization is irrecoverable for $\mathcal{O}$, we have to make a balance between the two networks such that the loss that occurs because of $\mathcal{P}$ can be recovered by $\mathcal{O}$. To solve this problem, we use alternating optimization; we give an equal chance to  $\mathcal{O}$ to recover from the loss. Hence $\mathcal{P}$ and $\mathcal{O}$ networks are optimized by one epoch iteratively.

    Our model contains binary variables. To make it differentiable, we use the scaled sigmoid $1/(1+e^{-\alpha x})$ where $\alpha$ is a hyperparameter; we increase $\alpha$ after a few epochs once the weights produced by the output layer of $\mathcal{P}$ is uniformly distributed in [0,1]. The high $\alpha$ value pulls the weights close to 0/1. This helps to get the approximate Bernoulli distribution in the \emph{pruner network} output space. 
    
    \vspace{-0.7em}
    \subsection{Results}
    \subsubsection{VGG-16 on CIFAR-10}
    CIFAR-10 is a widely used benchmark dataset, consisting of 50000 RGB images for training and 10000 images for testing. Each image is of size $32\times32$. For the data augmentation, we used a horizontal flip and random crop. The VGG-16 for CIFAR-10 contains the same architecture as \cite{vgg2014very}; the only difference is that a single 512-dimensional layer is used in place of the fully connected layers. We follow the same settings as in \cite{li2016pruning}. For the base model, the network is trained for 120 epochs and has an error rate of 6.51\%. The result of the proposed approach is shown in Table~\ref{tab:vgg-16cifar}.
    
    \begin{table}[t]
        \centering
        \scalebox{0.85}{
            \addtolength{\tabcolsep}{3.7pt}
            \begin{tabular}{|l| c| c| c|} 
                \hline
                Method &  Error\% & FLOPs & Pruned Flop\% \\ [0.8ex]
                \hline\hline
                Li-pruned  \cite{li2016pruning} &  6.60 & $2.06\times 10^8$ &34.20\\
                SparseVD \cite{molchanov2017variational} &  7.20 & -- & 55.95\\
                SBP  \cite{neklyudov2017structured} &  7.50 & -- & 56.52\\
                SBPa \cite{neklyudov2017structured} &  9.00 & -- & 68.35\\
                \hline\hline
                
                \textbf{NN-1 (Ours)} &  \textbf{6.74}&  $\mathbf{6.44\times10^7}$ & \textbf{79.47} \\
                \textbf{NN-2 (Ours)} &  \textbf{7.14}&  $\mathbf{5.33\times10^7}$ & \textbf{83.00} \\
                \textbf{NN-3 (Ours)} &  \textbf{7.47}&  $\mathbf{4.11\times10^7}$ & \textbf{86.90} \\
                \hline
            \end{tabular}
        }
        \vspace{3pt}
        \caption{Pruning result on the VGG-16 over the CIFAR-10 dataset (the baseline accuracy is 93.49\%).}
        \label{tab:vgg-16cifar}
        \vspace{-1.4em}
    \end{table}
    The rate of model compression depends on the regularization constant. Three different compressed models (\textit{NN-1, NN-2}, and \textit{NN-3}) can be obtained by just varying the regularization constant value that controls how many zeros we want in the multitask network. We use  0.001, 0.002 and 0.005 sparse regularization constant values in the \emph{pruner network} to obtain \emph{NN-1, NN-2} and \emph{NN-3} compressed models, respectively. Training of the network $\mathcal{P}$ and $\mathcal{O}$ is done in an alternating fashion. $\mathcal{P}$ tries to minimize the number of filters/feature-maps in the main network, while $\mathcal{O}$ recovers the accuracy.
    
    Table~\ref{tab:vgg-16cifar} shows that the proposed approach has a high pruning rate while still maintaining accuracy. In Table~\ref{tab:vgg-16cifar}, we can see that SBP \cite{neklyudov2017structured} has $7.5\%$ error on the 56.52\% pruning while SBPa shows the 68.35\% pruning with the 9.0\% error. Our proposed approach has only 7.47\% error with a high pruning rate of 86.9\%.
    
    \vspace{-1em}
    \subsubsection{ResNet-56 on CIFAR-10}
    \vspace{-0.5em}
    Next, we experiment on ResNet-56 over the CIFAR-10 dataset. It contains three stages of convolutional layers. Each layer is connected by projection mapping and followed by the average pooling and one fully connected layer. We use the same architecture and settings as described in \cite{li2016pruning}. The same alternate optimization, as described in the previous section, is performed for maximizing the filter pruning or maximizing the zero weights produced by the \emph{pruner network}. The network $\mathcal{P}$ is trained with the scaled sigmoid. Initially, we use scale $\alpha=1$, and after 30 epoch, we change the scale to $\alpha$=30. This new scale forces the output space of the \emph{pruner network} to be close to 0/1. Therefore we do not have any significant accuracy drop after discarding the filters corresponding to zero weights.
    
    Table~\ref{tab:resnet56} shows that the proposed approach achieves high compression rates while also giving the lowest error rate. In particular, SFP \cite{he2018soft} has error 6.65\% with the 52.6\% of FLOPs pruning while the proposed approach shows the significantly better pruning 61.51\% with the 6.61\% error rate. 
    
    \begin{table}[t]
        \centering
        \scalebox{0.85}{
            \addtolength{\tabcolsep}{4pt}
            \begin{tabular}{|l| c| c| c|} 
                \hline
                Method  & Error\% & FLOPs & Pruned Flop \% \\ [0.8ex] 
                \hline\hline
                Li-A \cite{li2016pruning}  & 6.90 & $1.12\times 10^8$ &10.40\\
                Li-B \cite{li2016pruning} & 6.94 & $9.04\times 10^7$ & 27.60\\
                NISP \cite{yu2017nisp}  & 6.99 & -- & 43.61\\
                CP \cite{channelPruning17}  & 8.20 & -- & 50.00\\
                SFP \cite{he2018soft} & 6.65 & -- & 52.60\\
                 AMC \cite{He_2018_ECCV} & 8.10 & -- & 50.00\\
                \hline\hline
                
                \textbf{NN-1 (Ours)}  &\textbf{6.61} & $\mathbf{4.85\times 10^7}$&\textbf{61.51} \\
                \hline
            \end{tabular}
        }
        \vspace{3pt}
        \caption{Pruning result of ResNet-56 architecture over CIFAR-10 dataset (the baseline accuracy is 93.1\%).}\vspace{-1em}
        \label{tab:resnet56}
         \vspace{-0.5em}
    \end{table}

    \begin{table}[b]
        \centering
        \scalebox{0.81}{
            \addtolength{\tabcolsep}{-2pt}
            \begin{tabular}{|l| c| c| c| c|}
                \hline
                Method & Acc\%(Top-1) & Acc\%(Top-5) & FLOPs Pruned \% \\ [0.8ex] 
                \hline\hline
                Baseline &  71.50 & 90.10 & --\\
                RNP (3X)\cite{lin2017runtime} &  -- &  87.57 & 66.67\\
                ThiNet-Conv \cite{luo2018thinet} &  69.74 &  89.41 & 69.04\\
                RNP (4X)\cite{rao2018runtime} &  -- &  86.67 & 75.00\\
                CP 4x\cite{channelPruning17} &  -- &  88.90 & 75.00\\ \hline\hline
                \textbf{NN-1 (Ours)} & \textbf{70.31} & \textbf{89.71} & \textbf{75.00}\\
                \hline
            \end{tabular} }
            \vspace{3pt}
            \caption{Pruning results for the VGG-16 over ImageNet. Our approach has minimum accuracy drop as compared to state-of-art pruning approach. We use the result reported in MatConvNet: http://www.vlfeat.org/matconvnet/pretrained/.}\vspace{-1em}
            \label{tab:vgg16Imagenet}
             \vspace{-0.5em}
        \end{table}
        \vspace{-1em}
    \subsubsection{VGG-16 on ImageNet}
    We evaluate our approach over the large-scale ImageNet dataset \cite{imagenet2015} using the VGG-16 architecture. The same ResNet-56 alternative optimization technique is used for pruning VGG-16 networks. We train $\mathcal{P}$ network with scaled sigmoid at the output layer. We use scale $\alpha=1$ for an initial 10 epochs, and then we set $\alpha=30$ for the rest of the training schedules. In Table-\ref{tab:vgg16Imagenet}, we compare our result with various other pruning approaches. As shown in Table-\ref{tab:vgg16Imagenet}, our approach gives $75\%$ FLOPs pruning with $89.71\%$ top-5 accuracy. On the other hand, CP-4x \cite{channelPruning17} gives 75\% FLOPs pruning with only 88.9\% top-5 accuracy. 
    
    \vspace{-1em}
    \subsubsection{ResNet-50 on ImageNet}
\begin{figure}[t]
        \centering
        \includegraphics[scale=0.25]{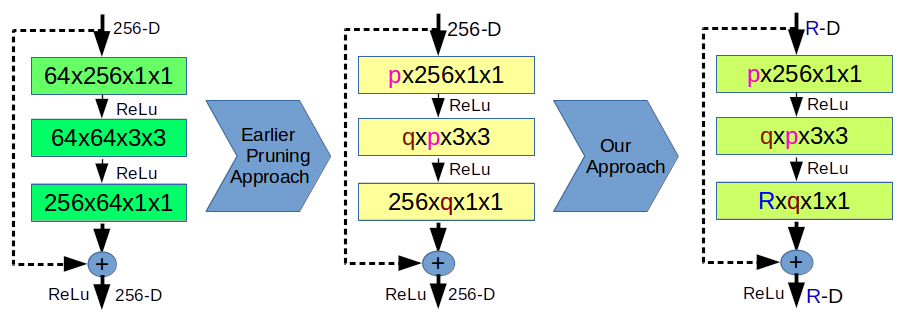}
        \caption{Unlike the previous approaches, our proposed method can also prune the skip connection filters. In the first two images, the skip connection size is fixed to 256-D, same as the original, while in the proposed approach, we also prune this to make it R-dimensional. }
        \label{fig:resnetpruning}
        \vspace{-12pt}
    \end{figure}

    ResNet-50 \cite{resnet} is a deep CNN architecture that has 50 layers with the residual connection. We use the same setup as proposed by the \cite{resnet}. 
    The previous approaches, such as \cite{yu2017nisp,he2018soft,ding2018auto}, etc., are unable to prune the skip connection filters because of the matrix addition inconsistency. These approaches only prune the middle layer filters, resulting in limited compression. In our approach, we also prune the skip connections. To solve the addition inconsistency, we give the same weights to the output filters and the skip connection filters. Therefore it prunes the same number of the filters in the output layers and the previous skip connection layers. Hence, the proposed approach can also prune the skip connection layers.  This may be very useful to prune complex networks, such as ResNet. Please refer to Figure \ref{fig:resnetpruning} for more details. 
    
    \begin{table}[t]
        \centering
        \scalebox{0.8}{
            \begin{tabular}{|l| c| c| c|} 
                \hline
                Method & Error\%(Top-1)  & Error\%(Top-5) &  Pruned Flop \% \\ [0.8ex] 
                \hline\hline
                Baseline & 24.7 & 7.8  & -\\
                ThiNet-70 \cite{luo2018thinet} & 25.97 & 7.9 & 36.8\\
                CP \cite{channelPruning17} & -- & 9.2 & $\sim$ 50\\
                NISP \cite{yu2017nisp} & 28.0 & -- & 44.0\\
                SFP \cite{he2018soft} & 25.39 & 7.94 & 41.8\\
                SPP \cite{wang2017structured} & -- & 9.6 & $\sim$ 50\\
                WAE \cite{chen2017learning} & -- & 9.6 & 46.8\\
                \hline\hline
                \textbf{NN-1 (Ours)} & \textbf{24.58} & \textbf{7.56} & \textbf{40.7}\\
                \textbf{NN-2 (Ours)} & \textbf{24.82} & \textbf{7.64} & \textbf{49.1}\\
                \hline
            \end{tabular}
        }
            \vspace{1pt}
        \caption{ResNet-50 Pruning results over the ImageNet dataset. The  accuracy  of  ResNet-50  is tested  using  official  1-crop  validation  setting:  center  224x224  crop  from  resized  image  with  shorter  side=256 (https://github.com/KaimingHe/deep-residual-networks).}
        \label{tab:resnet-50}
        \vspace{-1em}
    \end{table}
    
    In ResNet-50 Pruning, the \emph{pruner} is the multitask network with the 50 tasks, because of the 50 layers in the \emph{main-network}. We optimize the model in an alternating fashion for $\mathcal{P}$ and $\mathcal{O}$. In the first round, only $\mathcal{P}$ is optimized, while the parameters of $\mathcal{O}$ are kept frozen. In the second round, $\mathcal{P}$ and $\mathcal{O}$ is optimized jointly. The output dimension of each multitask output layer is equal to the number of filters in that layer. To get the output close to 0/1, scaled sigmoid is used. Initially, we set $\alpha=1$ for 10 epochs, and later we use $\alpha=50$ to get the Bernoulli weights (outputs of the multitask pruner network) on the feature maps. 
    
    Empirically, we found that our approach yields compressed ResNet-50  models (NN-1, NN-2) having significantly better accuracy as compared to other approaches \cite{yu2017nisp,channelPruning17,chen2017learning} because of skip connections pruning support. The proposed approach gives a significantly better pruning rate with the negligible accuracy drop. In Table \ref{tab:resnet-50}, we show a detailed comparison with other baselines.

        \begin{table*}[t]
            \centering
            \scalebox{.9}{
                \addtolength{\tabcolsep}{29pt}
                \begin{tabular}{|c|c|c|c|c|}
                    \hline
                    \multirow{2}{*}{\textbf{Model}}& \multirow{2}{*}{\textbf{data}} & \multicolumn{3}{c|}{\textbf{Avg. Precision, IoU:}}  \\ \cline{3-5}
                    & & \textbf{0.5:0.95} & \textbf{0.5} & \textbf{0.75}  \\ \hline
                    \textbf{F-RCNN original} & trainval35K & 30.3 & 51.3 & 31.8   \\ \hline
                    \textbf{F-RCNN pruned} & trainval35K & 30.2 & 51.0 & 31.6  \\ \hline
                \end{tabular}
            }
            \vspace{3pt}
            \caption{Generalization results over MS-COCO \cite{lin2014coco} dataset for Faster-RCNN object detector. In the original Faster-RCNN,  we use ResNet-50 as the base architecture while in the Faster-RCNN pruned, pruned ResNet-50 model (NN-2) from Table \ref{tab:resnet-50} is used.  We use a publicly available implementation (https://github.com/jwyang/faster-rcnn.pytorch) for Faster R-CNN with ResNet-50 as the base network.}
            \label{tab:coco}
             \vspace{-1.5em}
        \end{table*}
    
    \vspace{-0.5em}
        \subsection{Generalization}
            \vspace{-0.5em}
        To show the generalization ability of the compressed model produced by our proposed approach, we experiment on the object detection task. In this experiment, we select the popular Faster-RCNN \cite{ren2015fasterrcnn} architecture  on the large-scale MS-COCO \cite{lin2014coco} dataset. Our experimental results demonstrate that the compressed model produced by our proposed approach has the same generalization ability as the original model.

        \vspace{-1em}
        \subsubsection{Compression for Object Detection}
            \vspace{-0.5em}
        MS-COCO  \cite{lin2014coco} is a large-scale dataset, which contains 80 object categories. The training set contains 80K images, and the validation set contains 35K images in total; both are combined as used as the training set called trainval35K \cite{lin2017feature}. The object detection results are reported over the 5K unused validation images (minival). The Faster-RCNN \cite{ren2015fasterrcnn} is a highly popular object detection algorithm that takes the standard CNN as the base architecture for the feature extraction. For our experiments, we train the Faster-RCNN architecture with the ResNet-50 (uncompressed) \cite{resnet}  as the base network and the results are reported in Table \ref{tab:coco}. To show the generalization ability, we replace the base network ResNet-50 with the pruned ResNet-50 (NN-2) reported in Table \ref{tab:resnet-50}. Repeating the same procedure of the Faster-RCNN with the pruned base model, we achieve similar results, as shown in Table \ref{tab:coco}. Therefore our compressed model not only has high FLOPs saving but also better generalization ability and can be used to higher-level computer vision tasks. In the Faster-RCNN implementation, we use ROI Align and stride 1 for the last block of the convolutional layer (layer 4) in the base ResNet-50 model.  
            \vspace{-0.5em}
        \subsection{Practical Speedup}
        \begin{figure}[t]
            \centering
            \includegraphics[scale=0.38]{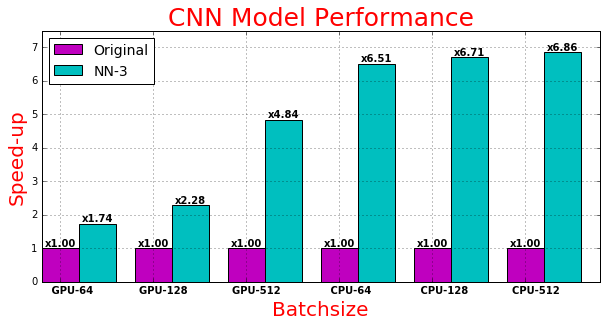}
            \caption{Practical speedup for the compressed model (NN-3) produced by the proposed approach (table-\ref{tab:vgg-16cifar}) w.r.t. batch size on VGG-16 architecture over CIFAR-10 dataset.}
            \label{fig:speedup}
             \vspace{-1.5em}
        \end{figure}
        
        In Fig-\ref{fig:speedup}, we demonstrate the practical speedup for VGG-16 compressed model (NN-3) given in the Table \ref{tab:vgg-16cifar}. As the Table shows, NN-3 compressed model has $7.63\times$ theoretical FLOPs compression. We achieve $4.84\times$, $6.86\times$ practical speedup corresponding to GPU and CPU with 512 batch size. Therefore, practical CPU speedup is close to the theoretical speedup, while the GPU's practical speedup is below the theoretical speedup. This is because of the availability of thousands of cores for computation in GPU. Here one can observe that, with the increase in batch size, the parallelization ability of the model also increases; therefore, the practical speedup is close to the theoretical FLOPs compression as shown in Fig-\ref{fig:speedup}.
        
        \vspace{-0.2em}
        \subsection{Ablation for Regularization Parameter}
            
        In Table-[\ref{tab:vgg-16cifar}, \ref{tab:resnet-50}] we conduct an ablation study over the $\lambda$ parameter mentioned in Eq.~\ref{eq:joint}. The $\lambda$ parameter is used to control the pruning rate in the model. If we increase the $\lambda$ value, it forces a high $l_1$ penalty to the multitask output vector and produces more zeros, while for lower values of $\lambda$, we get fewer zeros. These zero weight filters can be discarded from the model. In Table 1, NN1, NN2 and NN3 are compressed models for $\lambda=0.001, 0.002$ and $0.005$, respectively, and we achieve pruning rate 79.47\%, 83.00\% and 86.90\% respectively. Similarly, in table-4, NN1 and NN2 are compressed models for $\lambda=0.001$ and $0.002$, respectively. The detail compression rate and corresponding  accuracy can be seen in table [\ref{tab:vgg-16cifar}, \ref{tab:resnet-50}]. If we use too high pruning rate, it can dominate the model by discarding a large number of filters, and the model is unable to recover the performance. 
        
           \vspace{-0.5em}
        \section{Conclusion}
        We presented a filter pruning approach based on a multitask pruner network. The multitask network learns \textit{where to prune} in the main network. Alternating optimization used in the proposed approach helps to achieve high FLOPs pruning rate. The multitask network tries to maximize the pruning while the main network tries to maintain accuracy during pruning. The multitask network gives approximate Bernoulli weights to each feature map in the main-network and tries to maximize the number of such zero weights. Feature maps corresponding to the zero weights produce zero-valued feature maps in the output layer; therefore, these feature maps have no contribution in the overall model. We can safely remove these feature maps with corresponding filters from the main network without degrading the model performance. One of the appealing aspects of the proposed approach is that it can automatically decide the layer importance (where to prune). The proposed approach is end-to-end without any heuristics, such as an ad-hoc specification of thresholds for filter removal. The proposed approach yields state-of-art FLOPS pruning results with minimal accuracy drop and also shows a good generalization ability for the object detection task.

\textbf{Acknowledgment:} PS is supported by the Research-I Foundation at IIT Kanpur. VKV acknowledges support from Visvesvaraya PhD Fellowship and PR acknowledges support from Visvesvaraya Young Faculty Fellowship. 
        
        \newpage
        {\small
            \bibliographystyle{ieee}
            \bibliography{egbib}
        }
\end{document}